\documentclass[10pt,twocolumn,letterpaper]{article}

\usepackage{cvpr}              
\usepackage[accsupp]{axessibility}  

\usepackage{tummath}

\usepackage{csquotes}

\usepackage[
  acronym,
  automake,
]{glossaries-extra}
\makeglossaries

\setabbreviationstyle[acronym]{long-short-user}
\renewcommand*{\glsxtruserparen}[2]{
  \glsxtrfullsep{#2}%
  \glsxtrparen
   {#1\ifglshasfield{\glsxtruserfield}{#2}{;\xspace%
     \expandafter\citealp\expandafter{\glscurrentfieldvalue}%
   }{}%
   }%
}

\glsdefpostdesc{acronym}{
 \ifglshasfield{\glsxtruserfield}{\glscurrententrylabel}%
 {~\expandafter\citep\expandafter{\glscurrentfieldvalue}}%
 {}%
}

\glsdisablehyper
\defglsentryfmt[acronym]{\ifglsused{\glslabel}{\glsgenentryfmt}{\emph{\glsgenentryfmt}}}

\newacronym{gl:wsrd}{WSRD}{Würzburg shadow removal dataset}
\newacronym{gl:wsrdp}{WSRD+}{Würzburg shadow removal dataset plus}
\newacronym{gl:psnr}{PSNR}{peak signal-to-noise ratio}
\newacronym{gl:ssim}{SSIM}{structural similarity index}
\newacronym{gl:lpips}{LPIPS}{learned perceptual image patch similarity}
\newacronym{gl:fid}{FID}{Fréchet inception distance}
\newacronym{gl:istdp}{ISTD+}{Refined Image Shadow Triplets Dataset}
\newacronym{gl:UAV-SC}{UAV-SC+}{Refined Uncrewed Aerial Vehicle Dataset for Shadow Correction}

\usepackage{multirow}
\newcolumntype{x}{l}
\newcolumntype{X}{>{\scriptsize}l}
\newcolumntype{v}[1]{>{\raggedright\arraybackslash\hspace{0pt}}p{#1}}
\newcolumntype{V}[1]{>{\scriptsize\raggedright\arraybackslash\hspace{0pt}}p{#1}}

\definecolor{cvprblue}{rgb}{0.21,0.49,0.74}
\usepackage[pagebackref,breaklinks,colorlinks,allcolors=cvprblue]{hyperref}
\usepackage{xfrac}

\def\paperID{134} 
\def\confName{CVPR}
\def\confYear{2026}

\title{Winner of CVPR2026 NTIRE Challenge on Image Shadow Removal:\\Semantic and Geometric Guidance for Shadow Removal via Cascaded Refinement\thanks{Accepted at the CVPR 2026 Workshops (NTIRE 2026 Image Shadow Removal Challenge).}}  

\author{
Lorenzo Beltrame$^{1,2}$ \quad
Jules Salzinger$^1$ \quad
Filip Svoboda$^3$ \quad
Jasmin Lampert$^1$ \\ 
Phillipp Fanta-Jende$^1$  \quad
Radu Timofte$^4$ \quad
Marco K\"orner$^2$\\
$^1$Austrian Institute of Technology \quad
$^2$Technical University of Munich \quad
$^3$University of Cambridge \\ \quad
$^4$University of W\"urzburg\\
{\tt\small \{jules.salzinger, jasmin.lampert, phillipp.fanta-jende\}@ait.ac.at}\\
{\tt\small \{lorenzo.beltrame, marco.koerner\}@tum.de \quad}
{\tt\small fs437@cam.ac.uk \quad radu.timofte@uni-wuerzburg.de}
}

\begin{document}
\maketitle
\begin{abstract}
We present a three-stage progressive shadow-removal pipeline for the CVPR2026 NTIRE WSRD+ challenge. Built on OmniSR, our method treats deshadowing as iterative direct refinement, where later stages correct residual artefacts left by earlier predictions. The model combines RGB appearance with frozen DINOv2 semantic guidance and geometric cues from monocular depth and surface normals, reused across all stages. To stabilise multi-stage optimisation, we introduce a contraction-constrained objective that encourages non-increasing reconstruction error across the cascade. A staged training pipeline transfers from earlier WSRD pretraining to WSRD+ supervision and final WSRD+ 2026 adaptation with cosine-annealed checkpoint ensembling. On the official WSRD+ 2026 hidden test set, our final ensemble achieved 26.680 PSNR, 0.8740 SSIM, 0.0578 LPIPS, and 26.135 FID, ranked first overall, and won the NTIRE 2026 Image Shadow Removal Challenge. The strong performance of the proposed model is further validated on the ISTD+ and UAV-SC+ datasets.
\end{abstract}
\section{Introduction}
\label{sec:intro}

\begin{figure}[t]
    \centering
    \begin{subfigure}[t]{0.48\columnwidth}
        \centering
        \includegraphics[width=\linewidth]{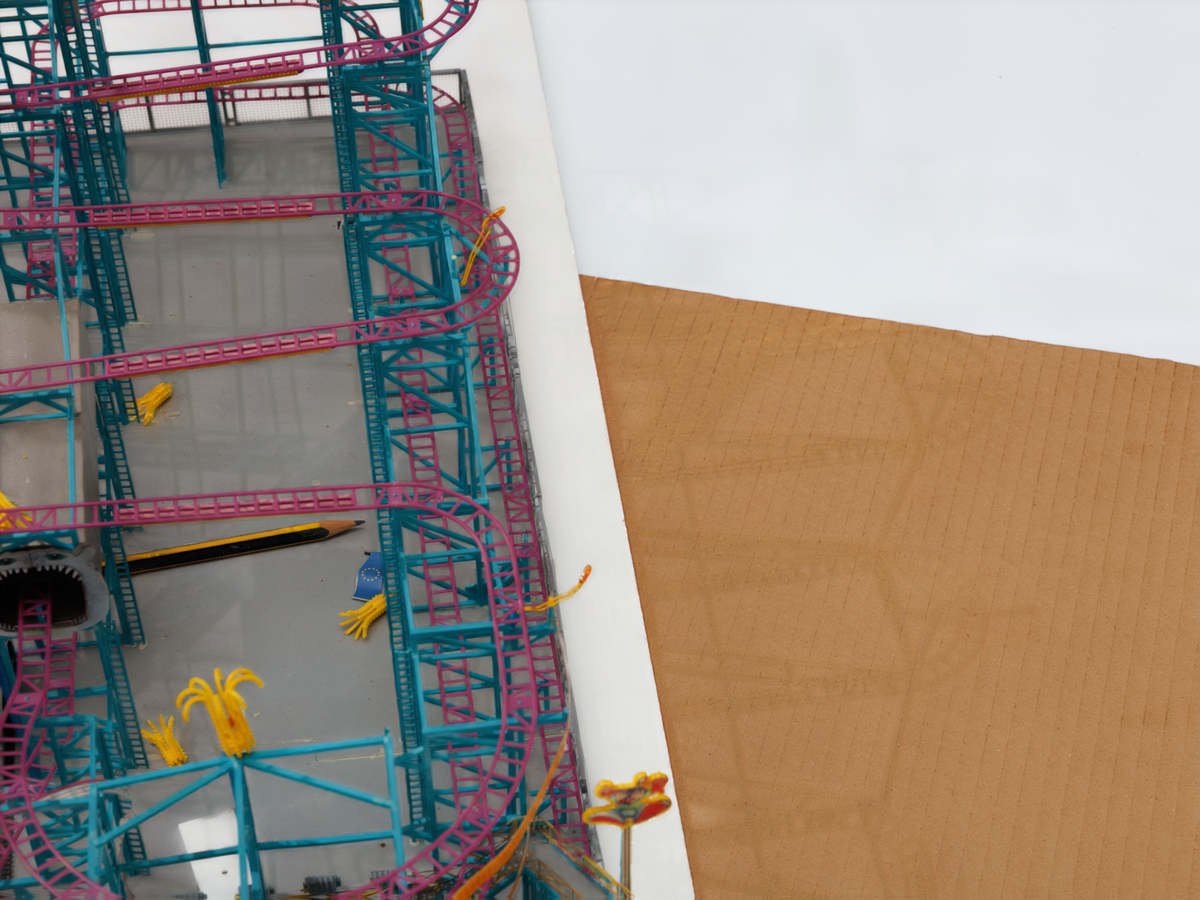}
        \caption{Single-stage OmniSR}
        \label{fig:intro_multi_a}
    \end{subfigure}
    \hfill
    \begin{subfigure}[t]{0.48\columnwidth}
        \centering
        \includegraphics[width=\linewidth]{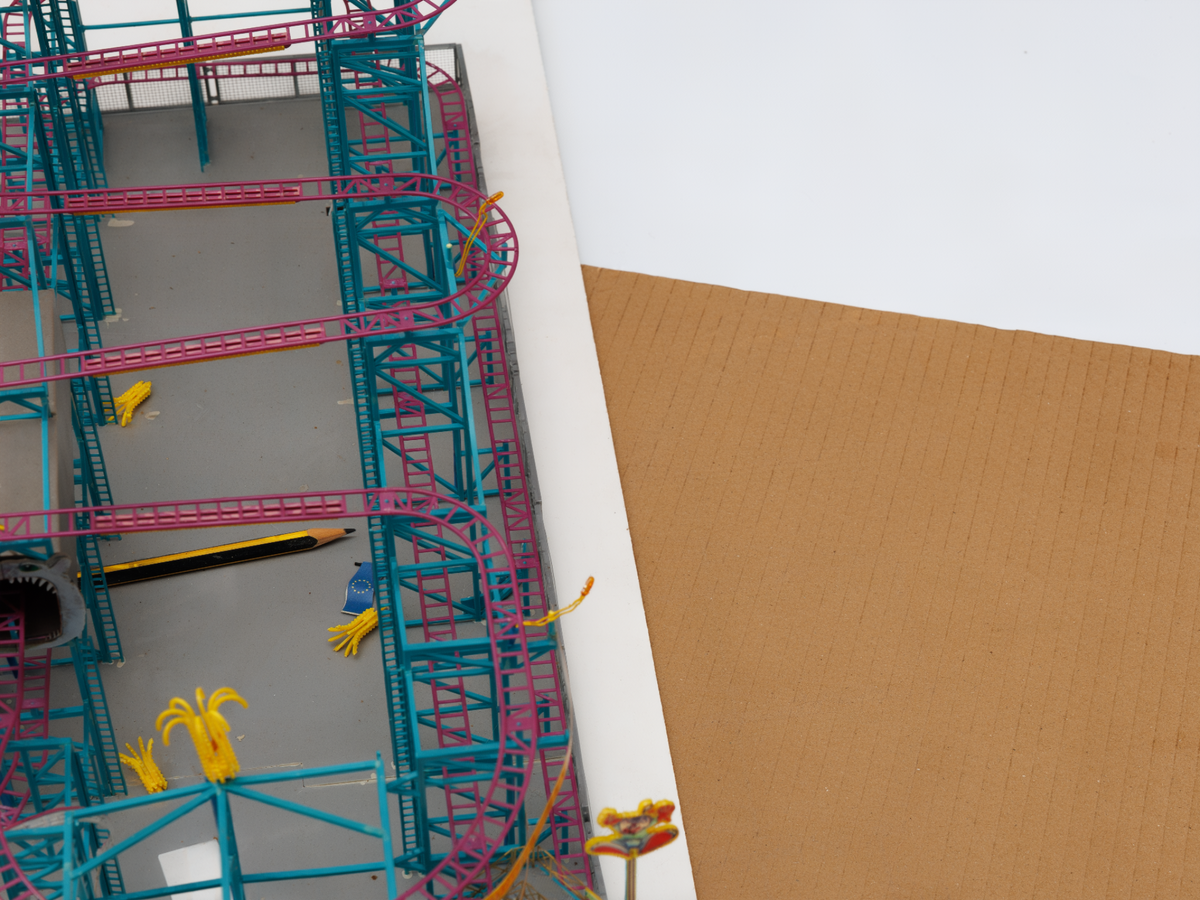}
        \caption{Three-stage OmniSR}
        \label{fig:intro_multi_b}
    \end{subfigure}

    \vspace{2pt}

    \begin{subfigure}[t]{0.48\columnwidth}
        \centering
        \includegraphics[width=\linewidth]{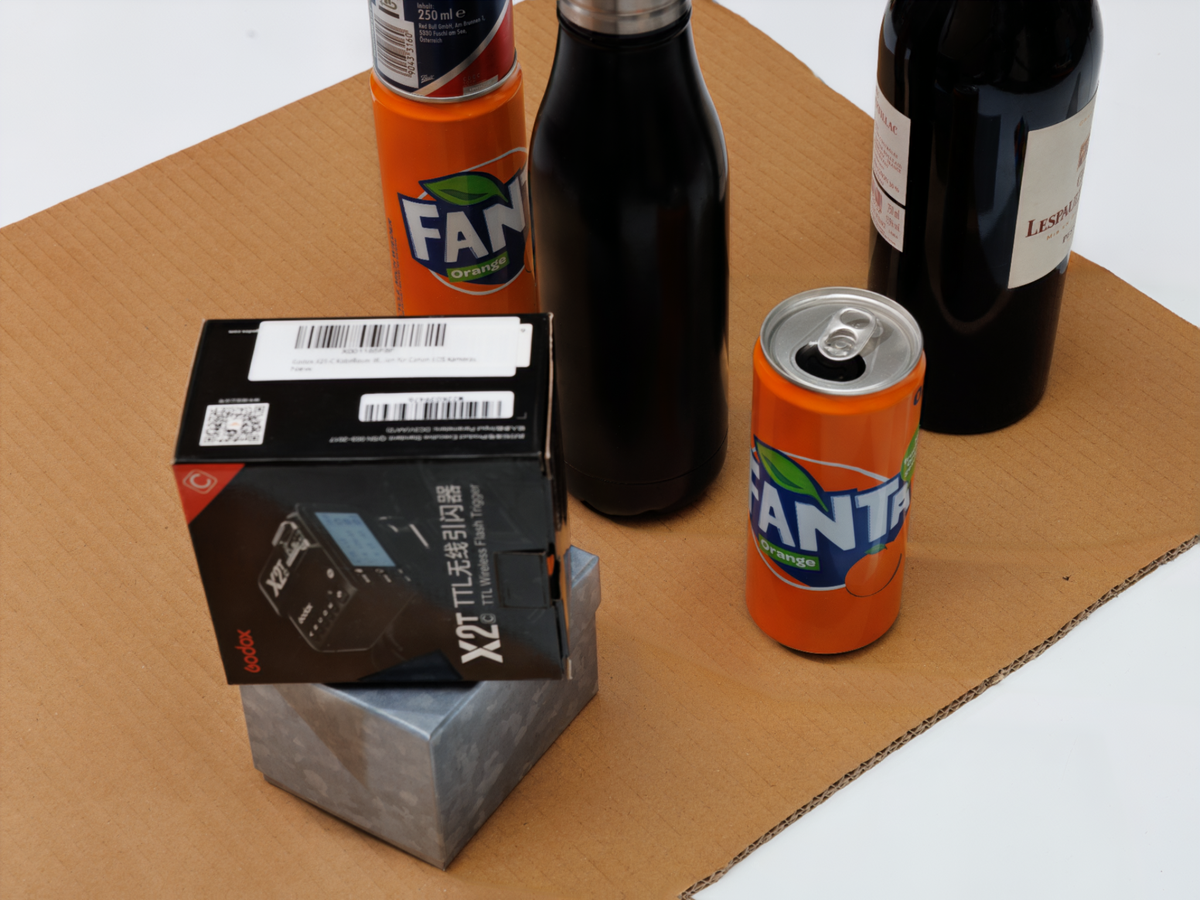}
        \caption{Single-stage OmniSR}
        \label{fig:intro_multi_c}
    \end{subfigure}
    \hfill
    \begin{subfigure}[t]{0.48\columnwidth}
        \centering
        \includegraphics[width=\linewidth]{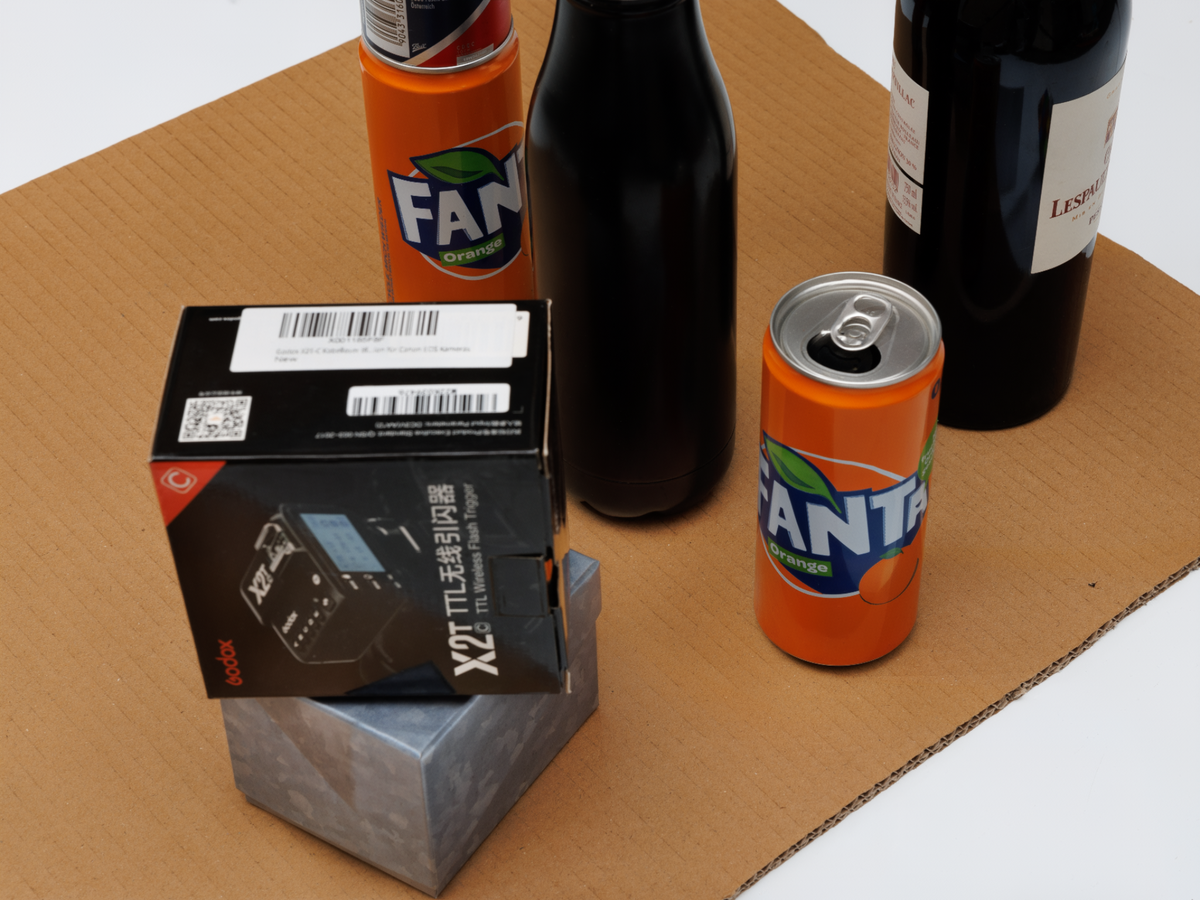}
        \caption{Three-stage OmniSR}
        \label{fig:intro_multi_d}
    \end{subfigure}
    \caption{Qualitative comparison on validation samples from WSRD+. The single-stage OmniSR predictions in \subref{fig:intro_multi_a} and \subref{fig:intro_multi_c} retain noticeable residual errors, including mild colour and illumination bias as well as local boundary artefacts. In contrast, the corresponding three-stage OmniSR predictions in \subref{fig:intro_multi_b} and \subref{fig:intro_multi_d} produce more consistent restoration with cleaner structures, supporting the use of progressive refinement.}
    \label{fig:intro_multistage}
\end{figure}

Image shadow removal is an important low-level vision problem because shadows corrupt local illumination, suppress visible texture, and reduce the reliability of downstream perception systems. 
This affects not only generic recognition and scene understanding, but also applications such as video analysis, traffic monitoring, and remote sensing~\cite{zhang2018improving,cucchiara2003detecting,arif2022comprehensive,alavipanah2022shadow}. 
In natural images, deshadowing is challenging because the model must separate illumination changes from intrinsic scene appearance while preserving texture, geometry, and object boundaries.

Recent deshadowing methods increasingly combine transformer-based restoration with auxiliary cues beyond RGB~\cite{wang2018stcgan,zhu2022bmnet,guo2023shadowdiffusion,guo2023shadowformer,xiao2024homoformer}, yet even strong single-stage systems such as OmniSR~\cite{xu2025omnisr} may still retain residual colour shifts, illumination bias, and boundary artefacts after one forward pass.
As shown in \cref{fig:intro_multistage}, the single-stage OmniSR results in \subref{fig:intro_multi_a} and \subref{fig:intro_multi_c} exhibit visible residual errors on WSRD+ validation samples~\cite{vasluianu2023wsrd}, whereas the corresponding three-stage OmniSR results in \subref{fig:intro_multi_b} and \subref{fig:intro_multi_d} yield a more uniform appearance and cleaner local transitions.

These limitations after a single forward pass suggest that shadow removal is better viewed as progressive refinement rather than one-shot prediction, consistent with iterative restoration formulations in inverse problems~\cite{venkatakrishnan2013plug,romano2017red,reehorst2019red,ryu2019plug,tan2025denoisers_inverse_problems}. On this basis, our method extends OmniSR to a three-stage direct-refinement architecture. To stabilise optimisation, we use a contraction-constrained objective that encourages monotonic improvement across stages, as detailed in \cref{sec:training_objective}. We further adopt staged pretraining across \gls{gl:wsrd}, \gls{gl:wsrdp}, and \gls{gl:wsrdp} 2026~\cite{vasluianu2023wsrd}. Additionally, we also evaluate on the \gls{gl:istdp}~\cite{wang2018stcgan} and \gls{gl:UAV-SC}~\cite{luo2023evolutionary} (\cref{tab:istdplus_mask_usage_unified} and \cref{tab:uavsc-whole}).

Our contributions are as follows:
\begin{itemize}
    \item We introduce a three-stage OmniSR-based shadow-removal pipeline for progressive direct refinement, coupled with contraction-constrained multi-stage supervision that stabilises training and promotes monotonic improvement across stages.
    \item We combine RGB appearance with DINOv2 semantic guidance and depth-derived geometric cues for robust deshadowing under challenging illumination conditions in a multi-stage setting.
    \item We present a staged pretraining regime for deshadowing that transfers from earlier imperfectly aligned WSRD data to aligned WSRD+ data and then adapts to the WSRD+ 2026 distribution, yielding consistent gains under dataset shift.
\end{itemize}

\glsresetall
\section{Related Work}
\label{sec:related}

Shadow removal has progressed from hand-crafted illumination models and decomposition approaches to fully learned image-to-image restoration systems. Early work studied the effect of shadows on recognition and scene understanding~\cite{zhang2018improving,cucchiara2003detecting}, while later deep methods learned shadow-specific priors directly from paired or synthetic data~\cite{inoue2020learning,le2019shadow}. Auxiliary guidance has remained a recurring design choice in this literature. ST-CGAN jointly predicts a shadow mask and a shadow-free image~\cite{wang2018stcgan}, BMNet conditions restoration on shadow masks and shadow-invariant color cues~\cite{zhu2022bmnet}, and ShadowDiffusion integrates degradation priors while refining masks during diffusion-based restoration~\cite{guo2023shadowdiffusion}. In parallel, transformer-based restoration frameworks have improved the ability to model long-range structure and non-local illumination effects, as illustrated by ShadowFormer and HomoFormer~\cite{guo2023shadowformer,xiao2024homoformer}. Among these methods, OmniSR~\cite{xu2025omnisr} is particularly relevant because it explicitly combines RGB appearance with semantic and geometric guidance for shadow removal under both direct and indirect lighting. These two characteristics are considered crucial in the shadow removal community~\cite{vasluianu2024ntire}.

\subsection{Semantic and geometric guidance}

Shadow removal is ambiguous when appearance alone cannot distinguish cast shadows from intrinsically dark materials or low-reflectance surfaces. Earlier auxiliary signals in deshadowing were typically mask- or prior-based, including detection outputs, decomposition cues, colour guidance, and degradation models~\cite{wang2018stcgan,le2019shadow,zhu2022bmnet,guo2023shadowdiffusion}. More recent work has begun to inject higher-level scene representations directly into the restoration backbone. DINOv2~\cite{oquab2023dinov2} supplies robust semantic representations without task-specific supervision, while monocular depth estimation with Depth Anything V2~\cite{depth_anything_v2} provides geometric structure from which surface normals can be derived. OmniSR~\cite{xu2025omnisr} demonstrated the utility of this specific semantic-plus-geometry conditioning scheme for shadow removal, and our method builds directly on that observation.

\subsection{Iterative and progressive restoration}

Many restoration problems benefit from repeated refinement rather than one-shot prediction. This perspective is well established in inverse problems, plug-and-play reconstruction, and regularisation-by-denoising frameworks, where learned priors are repeatedly applied to improve reconstruction fidelity~\cite{venkatakrishnan2013plug,romano2017red,reehorst2019red,ryu2019plug,tan2025denoisers_inverse_problems}. We adopt this view for shadow removal and formulate the proposed system as a multi-stage direct-refinement pipeline in which later stages explicitly correct the residual errors left by earlier ones.

\section{Method}
\label{sec:method}

\begin{figure*}[http]
    \centering
    \includegraphics[width=0.9 \textwidth]{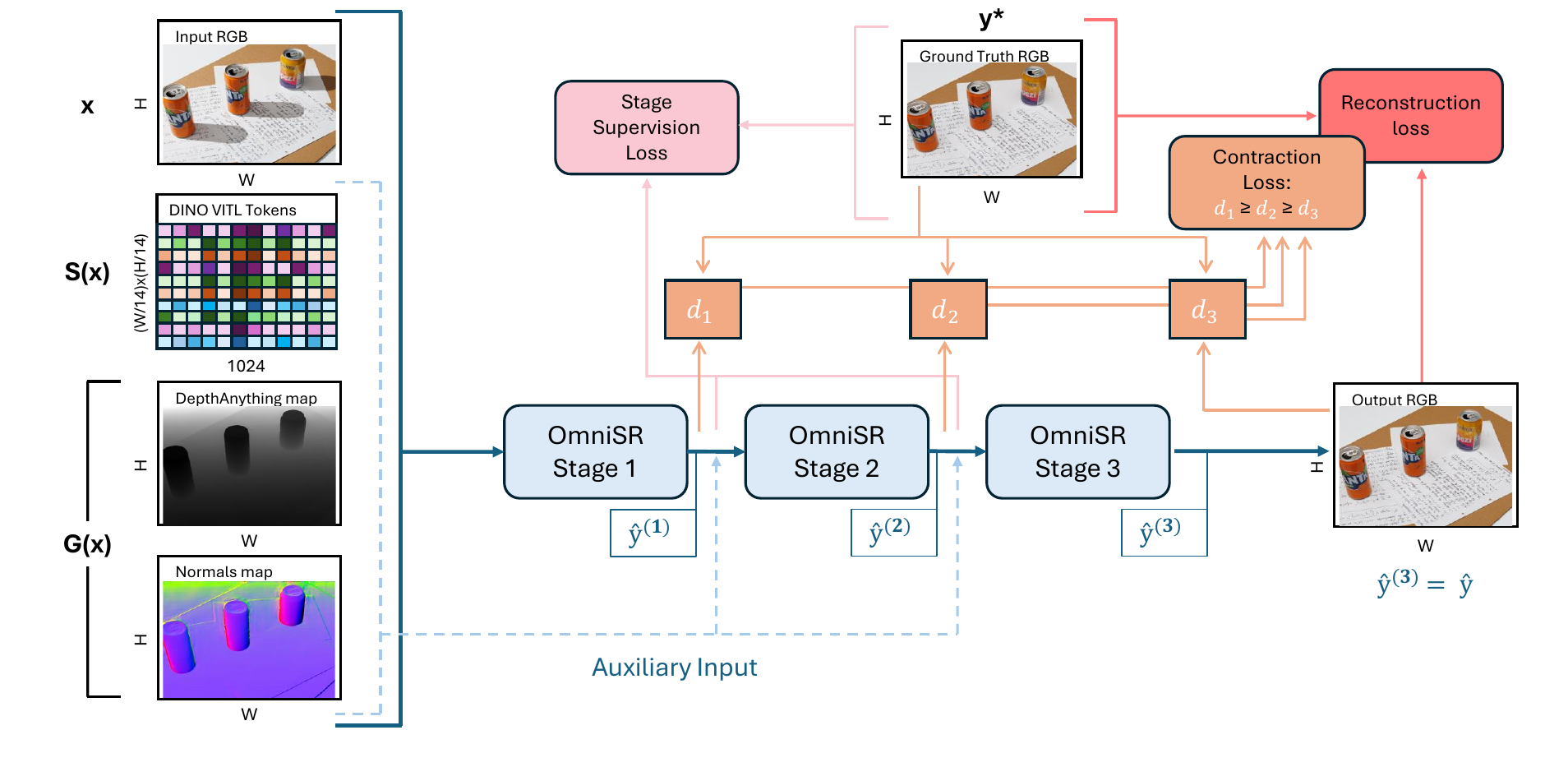}
    \caption{
    Proposed three-stage OmniSR pipeline. 
    The shadowed RGB input $\V{x}$ is processed by three cascaded OmniSR stages, where stage $k$ takes 
    $\hat{\V{y}}^{(k-1)}$ as input and outputs a refined prediction $\hat{\V{y}}^{(k)}$. 
    DINOv2 semantic features and depth-derived geometric cues are extracted once from $\V{x}$ and reused across all stages. 
    The final prediction is $\hat{\V{y}}^{(3)}$. 
    Training supervision is shown in the same diagram for compactness: 
    $d_k=\norm{\hat{\V{y}}^{(k)}-\V{y}^{*}}$ denotes the stage-wise error to ground truth, 
    the stage reconstruction loss ($\mathcal{L}_{\text{stage}}$) supervises the intermediate outputs, the reconstruction loss constrains the final output, 
    and the contraction loss encourages non-increasing error across stages.
    }
    \label{fig:multistage_omnisr}
\end{figure*}

Given a shadowed RGB image $\V{x} \in \R^{H \times W \times 3}$, the goal is to predict 
a reconstruction $\hat{\V{y}} \approx \V{y}^{*}$ as similar as possible to the ground-truth shadow-free image $\V{y}^{*}$.
Our proposed approach adopts a three-stage transformer-based iterative restoration cascade built on OmniSR~\cite{xu2025omnisr}.

Let $S(\V{x})$ denote frozen \emph{semantic} features extracted once from the input image using DINOv2~\cite{oquab2023dinov2}, and let $G(\V{x})$ denote \emph{geometric} guidance derived from monocular depth and surface normals estimated from the same input~\cite{depth_anything_v2}. 
Let further $f_k(\cdot)$ denote the restoration function implemented by the $k$-th OmniSR stage, and let  
\begin{align}
    \hat{\V{y}}^{(k)} &= f_k\left( \V{y}^{(k-1)}, S(\V{x}), G(\V{x}) \right)
\end{align}
denote the restored image predicted after stage $1 < k \leq K$
starting with $\hat{\V{y}}^{(1)} = f_1\left( \V{x}, S(\V{x}), G(\V{x}) \right)$.
The final prediction is taken as the output of the last stage, \ie, $\hat{\V{y}}=\hat{\V{y}}^{(K)}$.

The overall system, therefore, follows a direct multi-stage chaining strategy:
after the first stage maps the shadowed input $\V{x}$ to an initial restoration, and each subsequent stage takes the previous prediction $\hat{\V{y}}^{(k-1)}$ as its image input and produces a refined output $\hat{\V{y}}^{(k)}$. 
Within each stage, the OmniSR backbone retains its residual restoration design, predicting a correction relative to its own image input. 
The cascade implements an iterative residual correction process, where the first stage produces the main shadow-removal update, while later stages apply smaller refinements to the previous prediction, improving residual tonal inconsistencies and local artefacts.


\subsection{Semantic and geometric guidance}

Each stage combines RGB appearance with fixed semantic and geometric cues extracted from the original shadowed input. 

We use a frozen DINOv2 ViT-L/14 encoder~\cite{oquab2023dinov2} to extract semantic features from the \emph{shadowed} input image. 
Four intermediate DINO feature maps are obtained and projected to the feature width expected by the restoration network. 
Their concatenation is fused at the bottleneck, while the deepest projected DINO feature map is resized and injected into the deeper transformer blocks operating at $\sfrac{1}{4}$ and $\sfrac{1}{8}$ resolution. 
These DINO features are computed once from the original input and reused across all refinement stages to provide information about the images' semantic content, including shadows.

We estimate monocular depth using Depth Anything V2~\cite{depth_anything_v2} and derive geometric cues from it. 
The normalised depth channel $z$  is concatenated with the RGB image at the network input, producing an RGB-D representation. 
In addition, depth-derived point maps and surface normals are resampled to the internal working resolutions and injected into the deeper transformer blocks. 
This conditioning provides geometric structure during refinement and helps prevent appearance corrections from propagating across inconsistent scene regions.

\subsection{Progressive training and stage-wise expansion}\label{sec:pretraining_description}

Our final model is produced by a progressive training pipeline that combines robust pretraining on earlier data, stage-wise expansion of the cascade, and a final short adaptation run for model selection.

\paragraph{Phase 1: robust single-stage pretraining on earlier WSRD data.}
We begin by training a single-stage deshadowing model for 500 epochs on the earlier WSRD dataset from NTIRE 2023~\cite{Vasluianu_2023_CVPR}, in which the shadowed inputs and shadow-free targets are only approximately aligned. We retain the checkpoint with the highest validation PSNR, using a cosine-restart learning-rate schedule with 200-epoch cycles. We employ cosine warm restarts with a peak learning rate of \(1.0\times10^{-4}\), a minimum learning rate of \(5.0\times10^{-5}\), and a restart period of 200 epochs. At the beginning of each cycle, the learning rate is linearly warmed up for 5 epochs from \numrange[range-phrase={ to }]{e-5}{e-4}. 
The annealing range is intentionally kept narrow, since stronger decay was unnecessary under the restart scheduling regime.
Rather than treating the images' mismatch solely as a limitation, we use it as a source of a robust pretraining signal. 
Training under slight misalignment encourages the model to learn features that are tolerant to spatial shifts, boundary inaccuracies, and imperfect correspondence.

\paragraph{Phase 2: aligned two-stage transfer on WSRD+.}
We then transfer this pretrained initialisation to the aligned WSRD+ dataset used in NTIRE 2025~\cite{vasluianu2024ntire} and expand the model into a two-stage refinement system. This stage adapts the model to stronger supervision while preserving the robust deshadowing prior learned in Phase~1. We train this two-stage model for 1500 epochs, retain the checkpoint with the highest validation PSNR, and use the same cosine-annealing schedule as in the previous phase.

\paragraph{Phase 3: three-stage expansion and submitted configuration.}
Finally, we extend the aligned model into a multi-stage direct-refinement pipeline. The ablation results in \cref{tab:ablation_all} show that, under the available computational budget, $K=3$ achieves the best trade-off and is therefore used in the final configuration. The resulting three-stage model is initialised by transferring the first-stage weights from the first stage of the phase-2 model, while both the second and third stages are initialised from the residual-correction weights learned by the second stage in phase 2. This warm-start strategy equips the added stage with a meaningful residual-correction prior and stabilises optimisation of the full three-stage cascade.

\newcommand{\Kens}{\ensuremath{\Set{K}_\text{ens}}}%
\paragraph{Final adaptation and checkpoint averaging.}
After stage-wise expansion, we perform a final adaptation run on the WSRD+ 2026 training data for \num{100} epochs using cosine annealing and at a reduced learning rate. 
Rather than selecting a single terminal checkpoint, our final test-set prediction 
\begin{align}
    \hat{\V{y}}_{\text{ens}} &=
    \frac{1}{\abs{\Kens}} \sum_{k \in \Kens} f_{\V{\theta}_k}(\V{x})
\end{align} 
is obtained by averaging predictions from five checkpoints $k \in \Kens$ sampled along this adaptation trajectory.
In our experiments, we identified $\Kens = \SetDef{35,40,45,65,80}$, as 
epochs \numrange{35}{45} correspond to a plateau region on the older-distribution validation set, while epochs \numlist{65;80} occur later in the adaptation schedule and are expected to better match the 2026 data distribution. 
This temporal checkpoint ensemble provides mild output diversity at no additional training cost.

\subsection{Training objectives}
\label{sec:training_objective}

We trained the model in two phases, \ie an initial single-stage pretraining phase followed by joint optimization of the final multi-stage cascade. 

During single-stage pretraining, we optimise the combined loss
\begin{align}
    \mathcal{L}_{\text{pre}}
    &=
    \lambda_\text{MSE} \cdot
    \mathcal{L}_{\text{MSE}}
    +
    \lambda_\text{LPIPS} \cdot
    \mathcal{L}_{\text{LPIPS}}
\end{align}
consisting of 
the mean-squared reconstruction loss $\mathcal{L}_{\text{MSE}}$ and 
the perceptual similarity loss $\mathcal{L}_{\text{LPIPS}}$
computed from deep features~\cite{zhang2018lpips}.
The MSE term enforces pixel-level fidelity, while LPIPS encourages perceptual realism.

For the final multi-stage model, we optimize the overall objective
\begin{align}
    \mathcal{L}_{\text{total}} 
    &=
    \mathcal{L}_{\text{pre}} \\\notag
    &\quad 
    +
    \lambda_\text{hessian} \cdot
    \mathcal{L}_{\text{hessian}} 
    +
    \lambda_\text{stage} \cdot
    \mathcal{L}_{\text{stage}}  \\\notag
    &\quad
    +
    \lambda_\text{contraction} \cdot
    \mathcal{L}_{\text{contraction}} \quad
\end{align}
where $\mathcal{L}_{\text{pre}}$ serves as the primary reconstruction-perceptual objective, 
while $\mathcal{L}_{\text{hessian}}$, $\mathcal{L}_{\text{stage}}$, and $\mathcal{L}_{\text{contraction}}$ 
promote structural consistency, provide explicit supervision to intermediate stage predictions, and encourage progressive improvement across stages, respectively.

To quantify stage-wise reconstruction quality,
we define the error 
\begin{align}
    d_k
    &= 
    \norm{\V{y}^{(k)} - \V{y}^{*}}
\end{align}
at stage $k$ 
as the $\ell_2$ distance between the stage-$k$ prediction $\hat{\V{y}}^{(k)}$ and the ground-truth $\V{y}^{*}$ image.
%
This quantity is used to determine the stage supervision loss
\begin{align}
    \mathcal{L}_{\text{stage}}\left(\hat{\V{y}}^{(1:K-1)}, \V{y}^{*}\right)
    &= \frac{1}{2}
    \sum_{k=1}^{K-1} \left\| \hat{\V{y}}^{(k)} - \V{y}^{*} \right\|^2
    \label{eq:stage_loss}
\end{align}
by averaging the distances between the intermediate stage predictions $\hat{\V{y}}^{(1)}, \hat{\V{y}}^{(2)}, \ldots, \hat{\V{y}}^{(K-1)}$ and the ground-truth image $\V{y}^{*}$. 
This term provides direct supervision to the intermediate outputs, encouraging each stage to apply smaller residual corrections to obtain a shadow-free estimate, while the final stage is primarily governed by the pre-training loss $\mathcal{L}_{\text{pre}}$.

The contraction loss is then defined as
\begin{align}
    \mathcal{L}_{\text{contraction}}\left(\hat{\V{y}}^{(k)}, \V{y}^{*}\right) 
    &=
    \sum_{k=2}^{K}
    \left[
    d_k
    - 
    \operatorname{sg}\left(d_{k-1}
    \right)
    \right]_{+},
    \label{eq:contraction_loss}
\end{align}
where $[z]_{+}=\max(z,0)$ denotes the positive-part operator and $\operatorname{sg}(\cdot)$ denotes the stop-gradient operator. The stop-gradient operator ensures that the previous-stage error acts as a fixed reference during optimisation, so the contraction term only penalises increases in the current-stage error and does not backpropagate through earlier-stage predictions.
This term penalises only those stages whose reconstruction error exceeds that of the preceding stage, thereby encouraging a monotonically improving sequence of predictions without enforcing a fixed decay schedule. 
Together, these objectives promote accurate reconstruction, perceptual quality, structural consistency, and stable progressive refinement.

\subsection{Implementation details and evaluation protocol}

Our implementation is based on PyTorch and uses AdamW with cosine learning-rate decay.
In the submitted configuration, we use an initial learning rate of $\alpha_0=5\times10^{-5}$, a minimum learning rate of $\alpha_1=10^{-6}$, gradient clipping at $1.0$, random training crops of size $\qtyproduct{512 x 512}{\pixel}$, and batch size of $b_\text{train}=4$.
At inference time, we apply the same fixed semantic and geometric guidance to all three stages and return the output of the final stage.
For large validation images, we use mixed-precision tiled inference with a tile size of $s=896$ and an overlap of $o=128$.
For the controlled ablations, all variants share the same base architecture and training protocol.
Unless stated otherwise, all stages are jointly supervised with an intermediate-stage loss weight of $\lambda_{\mathrm{stage}}=0.5$, and a contraction loss with weight $\lambda_{\mathrm{contraction}}=1.0$.
The ablation models are trained with AdamW ($\beta_1=0.9$, $\beta_2=0.999$, weight decay $10^{-2}$) using cosine decay with $\eta_0=5\times10^{-5}$ and $\eta_{\mathrm{min}}=10^{-6}$.
The loss function combines an $\ell_2$ reconstruction term with LPIPS using a weight $\lambda=0.05$.
All ablation models are warm-started from a shared single-stage pretrained model, as described in \cref{sec:pretraining_description}, and are trained and evaluated on WSRD+. Due to GPU memory constraints, we use batch size 4 for $K\leq3$ and batch size 2 for $K\geq4$.

The training and evaluation protocol follows the progression of the benchmark datasets across challenge editions. WSRD and WSRD+ are used for pretraining, transfer, and model development: WSRD supports robust pretraining despite imperfect alignment, while WSRD+ provides better-aligned supervision for multi-stage transfer, validation, and model selection. WSRD+ 2026 is reserved for the final competition stage and used only for cosine-annealed fine-tuning and official test scoring, with test images inferred without tiling. Evaluation is conducted in three settings: the official result on the NTIRE 2026 WSRD+ hidden test set, progressive adaptation across WSRD, WSRD+, and WSRD+ 2026, and controlled ablations on the public WSRD+ validation split, including cascade depth and the main components of the final three-stage model under a shared pretrained initialization. For the ISTD+ and UAV-SC+, we rely on the protocol established in \cite{xu2025detail} and  \cite{chu2025rmmamba}, respectively. Unless otherwise stated, we report \gls{gl:psnr}, \gls{gl:ssim}~\cite{Wang04:IQA}, and \gls{gl:lpips}~\cite{zhang2018lpips}. For the official NTIRE 2026 benchmark submission, we additionally report \gls{gl:fid}~\cite{Jayasumana24:FID}.

\section{Experiments}
\label{sec:experiments}


\subsection{Main result on the NTIRE 2026 Shadow Removal Challenge}
\begin{table*}[t]
    \centering
    \small
    \setlength{\tabcolsep}{4pt}
    \caption{Public leaderboard results on the NTIRE 2026 Shadow Removal Challenge WSRD+ hidden test set~\cite{ntire2026shadow}. \enquote{Ours} corresponds to the ensemble submission. Runtime values are indicative only, since they were reported on different hardware platforms. }
    \label{tab:ntire2026-shadow-public-merged}
    \begin{tabular}{@{}Xccccccc>{\arraybackslash}c@{}}
        \toprule
        Method & \multicolumn{4}{X}{Metric} & \multicolumn{3}{X}{Efficiency} & \multicolumn{1}{X@{}}{Final Rank} \\
        \cmidrule(lr){2-5} \cmidrule(lr){6-8}
        ~ & \multicolumn{1}{X}{PSNR $\uparrow$} & \multicolumn{1}{X}{SSIM $\uparrow$} & \multicolumn{1}{X}{LPIPS $\downarrow$} & \multicolumn{1}{X}{FID $\downarrow$} & \multicolumn{1}{X}{Params. (M) $\downarrow$} & \multicolumn{1}{X}{Runtime (s) $\downarrow$} & \multicolumn{1}{X}{Device} & ~ \\
        \cmidrule(r){1-1}
        \cmidrule(lr){2-2} \cmidrule(lr){3-3} \cmidrule(lr){4-4} \cmidrule(lr){5-5}
        \cmidrule(lr){6-6} \cmidrule(lr){7-7} \cmidrule(lr){8-8}
        \cmidrule(l){9-9}
        \textbf{Ours} & \textbf{26.68} & \textbf{0.874} & \textbf{0.058} & \textbf{26.14} & 74.3 & 0.30 & H100 & \textbf{1} \\
        \addlinespace
        RAS & 26.14 & 0.866 & 0.071 & 30.47 & 1500 & 0.5 & A800 & 2 \\
        SNU-ISPL-B & 25.94 & 0.867 & 0.085 & 28.05 & 9.5 & 1.9 & RTX 3090 & 3 \\
        APRIL-AIGC & 26.45 & 0.848 & 0.079 & 29.65 & 9105 & 9.9 & H20 & 4 \\
        DiogenesCask & 26.43 & 0.868 & 0.086 & 33.64 & 40.8 & 2.5 & RTX 4090 & 5 \\
        SNUCV & 25.30 & 0.864 & 0.068 & 30.00 & 136 & 35 & A6000 & 6 \\
        Shadow Breaker & 25.64 & 0.864 & 0.085 & 33.47 & 24.8 & -- & A40 & 7 \\
        CV\_SVNIT & 25.59 & 0.858 & 0.079 & 33.23 & 379 & 1.0 & 2$\times$T4 & 8 \\
        SNU-ISPL-A & 25.82 & 0.865 & 0.088 & 36.47 & 29 & 3.1 & RTX 3090 & 9 \\
        LUMOS\_Shadow & 25.78 & 0.862 & 0.091 & 39.79 & 3.0 & 0.6 & A800 & 10 \\
        ULR & 26.04 & 0.838 & 0.116 & 38.17 & 3622 & 0.9 & H100 & 11 \\
        KLETech-CEVI & 24.68 & 0.842 & 0.117 & 35.74 & 406.8 & -- & -- & 12 \\
        PSU\_TEAM & 23.39 & 0.838 & 0.112 & 45.26 & 45.6 & 161 & A100 & 13 \\
        avovo & 23.23 & 0.827 & 0.099 & 50.31 & 117 & 1.3 & A100 & 14 \\
        \bottomrule
    \end{tabular}
\end{table*}

Our method ranked first in the NTIRE 2026 Shadow Removal Challenge~\cite{ntire2026shadow}. 
On the official WSRD+ 2026 hidden test set, our final cosine-annealed five-checkpoint ensemble obtained the best overall rank among all submissions. 
Our best ensemble is reported in Table~\ref{tab:ntire2026-shadow-public-merged}, the public leaderboard results. Moreover, our annealed checkpoint achieves  $\text{PSNR}= \qty{26.571}{\deci\bel}$,  $\text{SSIM} = \qty{0.8733}{}$, and  $\text{LPIPS} = \qty{0.0577}{}$ on WSRD+, while the bare pretrained model without fine-tuning achieves $\text{PSNR} = \qty{25.909}{\deci\bel}$, $\text{SSIM} = \qty{0.8616}{}$, and $\text{LPIPS} = \qty{0.0688}{}$. Note that the WSRD+ 2026 test set differs from the earlier WSRD+ benchmark, so the reported values are not directly comparable across these two distributions. 

Compared with the strongest competing solutions in the NTIRE2026 Challenge, our method occupies a distinct design point.
The second-ranked RAS and the fourth-ranked APRIL-AIGC both rely on large diffusion-based backbones, using a second stage primarily for fidelity recovery after a strong generative de-shadowing pass, whereas our approach remains fully restoration-oriented and performs explicit stage-wise residual correction throughout the cascade.
By contrast, the third-ranked SNU-ISPL-B and the fifth-ranked DiogenesCask are closer to our method in that they also exploit strong auxiliary priors such as DINO-based semantics and geometric cues; however, their emphasis is on feature fusion, cross-attention, and frequency-aware reconstruction within one- or two-stage architectures rather than on progressive multi-stage refinement.
Taken together, the leaderboard suggests that auxiliary semantic/geometric guidance is a common ingredient among the best-performing methods, but that our combination of such priors with a three-stage direct-refinement cascade, contraction-constrained supervision, and staged dataset adaptation provides a particularly effective balance of restoration fidelity and perceptual quality.


\subsection{Progressive adaptation across dataset versions}

\begin{table}[t]
    \centering
    \small
    \setlength{\tabcolsep}{4pt}
    \caption{Chronological training trajectory across dataset versions, evaluated on the public WSRD+ validation split.
    Each row corresponds to the best checkpoint after the respective training phase selected based on the highest PSNR.}
    \label{tab:progression_results}
    \begin{tabular}{@{}lcccc@{}}
            \toprule
            Training Dataset & Stages & PSNR $\uparrow$ & SSIM $\uparrow$ & LPIPS $\downarrow$ \\
            \midrule
            WSRD & 1 & 20.256 & 0.673 & 0.1566 \\
            WSRD+& 2 & 27.184 & \textbf{0.879} & \textbf{0.0610} \\
            WSRD+& 3 & \textbf{27.356} & 0.877 & 0.0631 \\
            WSRD+ 2026 & 3 & 27.255 & 0.876 & 0.0625 \\
            
            \bottomrule
        \end{tabular}%
\end{table}

We next summarise the cumulative training trajectory across WSRD, WSRD+, and WSRD+ 2026. All results in Table~\ref{tab:progression_results} are evaluated on the same public WSRD+ validation split, but should not be interpreted as a controlled ablation, since both the supervision regime and the number of refinement stages change across phases. Instead, the table reflects how the model evolves as it is progressively adapted from earlier, imperfectly aligned supervision to increasingly target-specific data.

The dominant improvement comes from transferring from WSRD 2023 to the better-aligned WSRD+ supervision. This suggests that the initial pretraining mainly provides a robust shadow-removal prior, while most of the measurable restoration gain is realised only after adaptation to better-aligned data. In contrast, the final adaptation to WSRD+ 2026 with three-stage refinement results in a slight performance decrease, which likely reflects a domain shift between WSRD+ and WSRD+ 2026, given that model selection and validation are conducted on WSRD+. At the same time, the perceptual metric does not improve monotonically, suggesting a mild trade-off in using PSNR as the selection metric between distortion-oriented fidelity and perceptual quality during the final adaptation step.

\subsection{Ablation study}
\begin{table}[t]
  \centering
  \small
  \setlength{\tabcolsep}{5pt}
  \caption{Ablation results on the public WSRD+ validation set. The upper block reports the effect of the number of cascade refinement stages (a), where each checkpoint is selected by the highest validation PSNR. The lower block reports component ablations under a shared pretrained initialisation and fixed 3-stage OmniSR architecture (b), training schedule, and dataset split.}
  \label{tab:ablation_all}
  \begin{tabular}{@{}llccc@{}}
    \toprule
     & Ablation variant & PSNR $\uparrow$ & SSIM $\uparrow$ & LPIPS $\downarrow$ \\
    \midrule
    \multirow{5}{*}{(a)}
      & 1 stage  & 27.077 & 0.873 & 0.0605 \\
      & 2 stages & 27.274 & 0.877 & \textbf{0.0599} \\
      & \textbf{3 stages} & \textbf{27.356} & 0.877 & 0.0631 \\
      & 4 stages & 27.265 & 0.877 & 0.0646 \\
      & 5 stages & 27.307 & \textbf{0.878} & 0.0662 \\
    \midrule
    \midrule
    \multirow{5}{*}{(b)}
      & Full model              & \textbf{27.356} & \textbf{0.877} & 0.0631 \\
      & w/o contraction loss    & 27.173 & \textbf{0.877} & \textbf{0.0608} \\
      & w/o DINOv2 guidance     & 25.859 & 0.871 & 0.0711 \\
      & w/o depth \& normal     & 27.105 & 0.876 & 0.0634 \\
      & w/o depth           & 27.253 & 0.877 & 0.0636 \\
    \bottomrule
  \end{tabular}
\end{table}

\paragraph{Effect of the number of cascade stages.}
\Cref{tab:ablation_all} reports the best-checkpoint performance on the validation split for $K \in \SetDef{1,2,3,4,5}$ stages. A single stage already provides competitive performance, and adding a second stage yields a modest gain. 
The three-stage model achieves the best \gls{gl:psnr} at \qty{27.356}{\deci\bel}, indicating that three cascaded refinements are sufficient to remove most residual shadow artefacts. Further increasing the depth does not improve \gls{gl:psnr}, suggesting diminishing returns at this dataset scale. We therefore select $K=3$ as the final configuration. While \gls{gl:ssim} is comparable for different numbers of stages, \gls{gl:lpips} is best for the two-stage model and gradually decreases for $K \geq 3$, suggesting that additional stages may over-smooth fine perceptual details despite small structural gains.

\paragraph{Effect of omitting components during fine-tuning.}
\Cref{tab:ablation_all} shows that the semantic guidance is the most important component during WSRD+ adaptation, as removing DINOv2 leads to the clearest degradation across all metrics. This result is consistent with the single-stage observation in \cite{xu2025omnisr}. The geometric cues provide a smaller but still consistent benefit, with the joint depth-and-normal conditioning giving the strongest overall results and the depth-only ablation suggesting that part of the structural information can be recovered from normals alone. By contrast, removing the contraction loss produces a mixed effect: distortion-oriented metrics deteriorate, while LPIPS improves slightly. This indicates that the contraction term mainly acts as a stabilising regulariser for progressive refinement, favouring faithful stage-wise correction over purely perceptual sharpness. Overall, the ablation supports the final design choice: semantic guidance provides the dominant gain, geometry contributes complementary structural information, and the contraction loss improves the stability and fidelity of the fine-tuned multi-stage model.

\subsection{Qualitative results on the WSRD+ 2026 test set}
\begin{figure}[t]
    \centering
    \setlength{\tabcolsep}{1.5pt}
    \renewcommand{\arraystretch}{0.9}
    \begin{subfigure}[t]{0.48\textwidth}
        \centering
        \begin{tabular}{@{}cc@{}}
            \scriptsize Input & \scriptsize Output \\
            \includegraphics[width=0.48\linewidth]{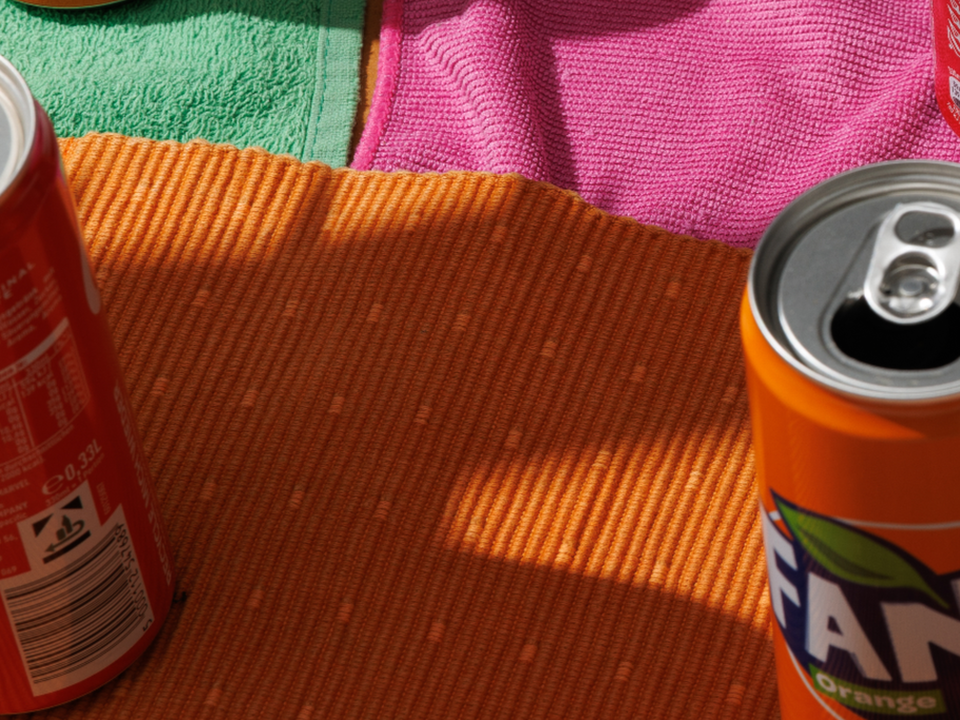} &
            \includegraphics[width=0.48\linewidth]{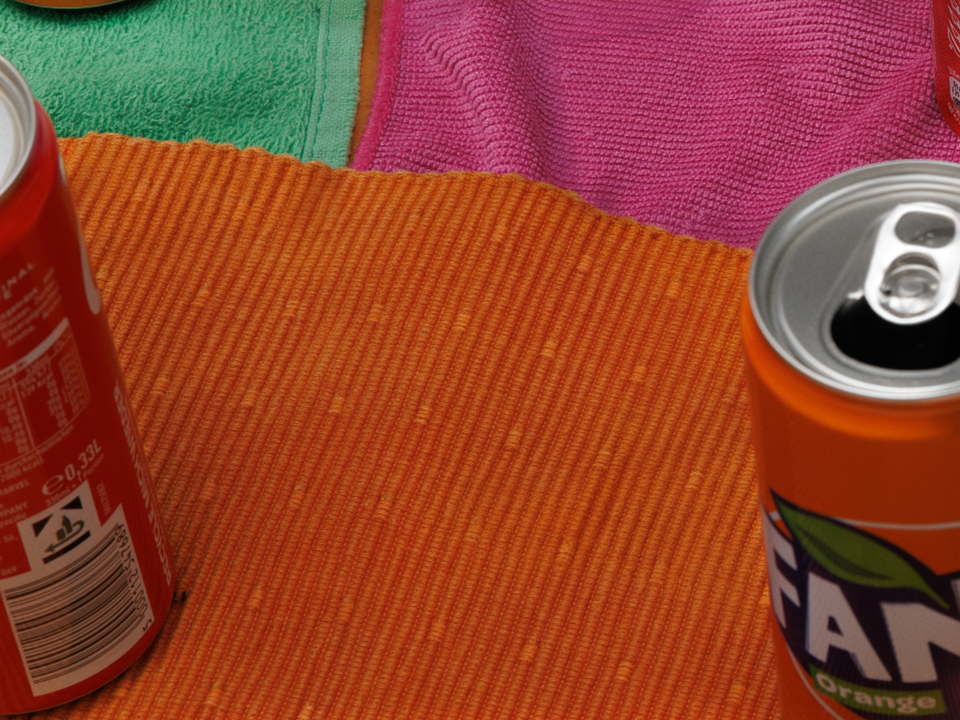}
        \end{tabular}
        \caption{Textured fabric and reflective cans.}
    \end{subfigure}

    \vspace{1mm}

    \begin{subfigure}[t]{0.48\textwidth}
        \centering
        \begin{tabular}{@{}cc@{}}
            \includegraphics[width=0.48\linewidth]{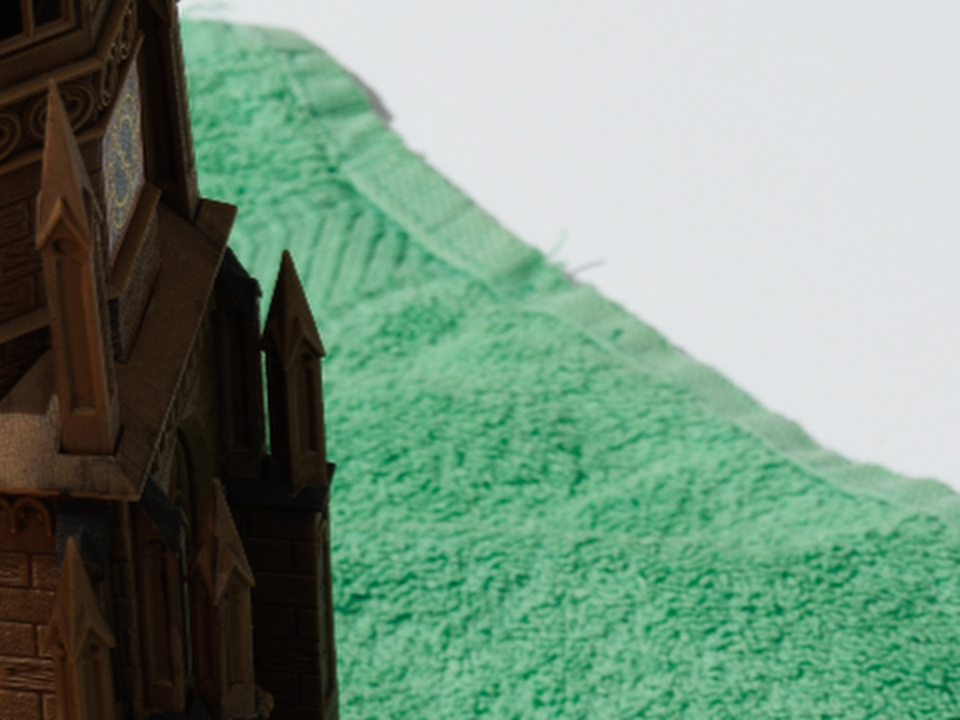} &
            \includegraphics[width=0.48\linewidth]{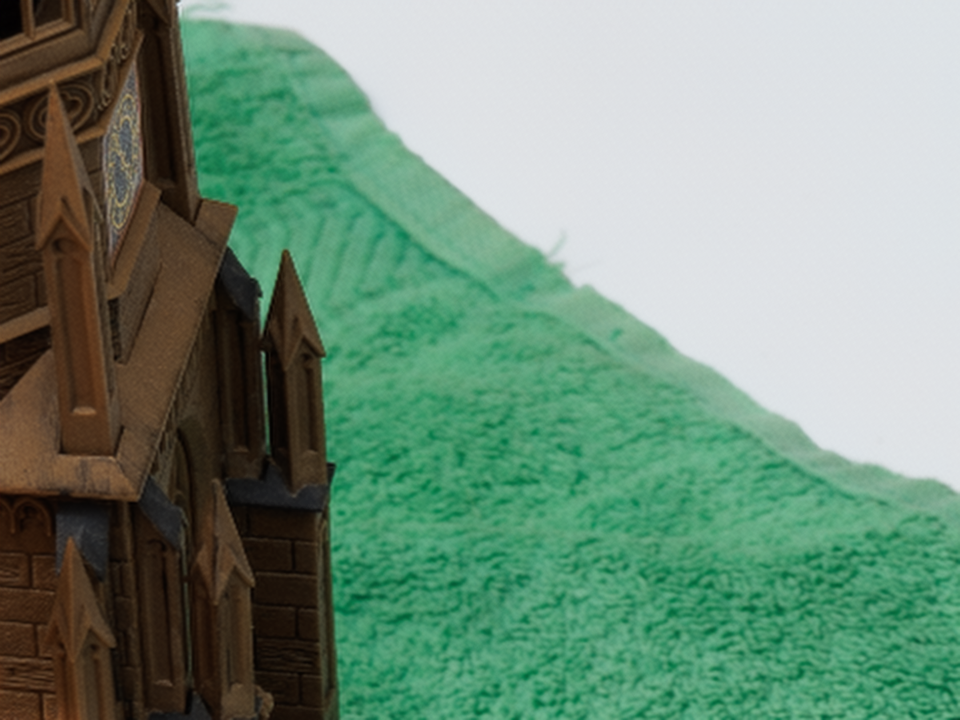}
        \end{tabular}
        \caption{Fine boundary structures.}
    \end{subfigure}\hfill

    \vspace{1mm}
        
    \begin{subfigure}[t]{0.48\textwidth}
        \centering
        \begin{tabular}{@{}cc@{}}
            \includegraphics[width=0.48\linewidth]{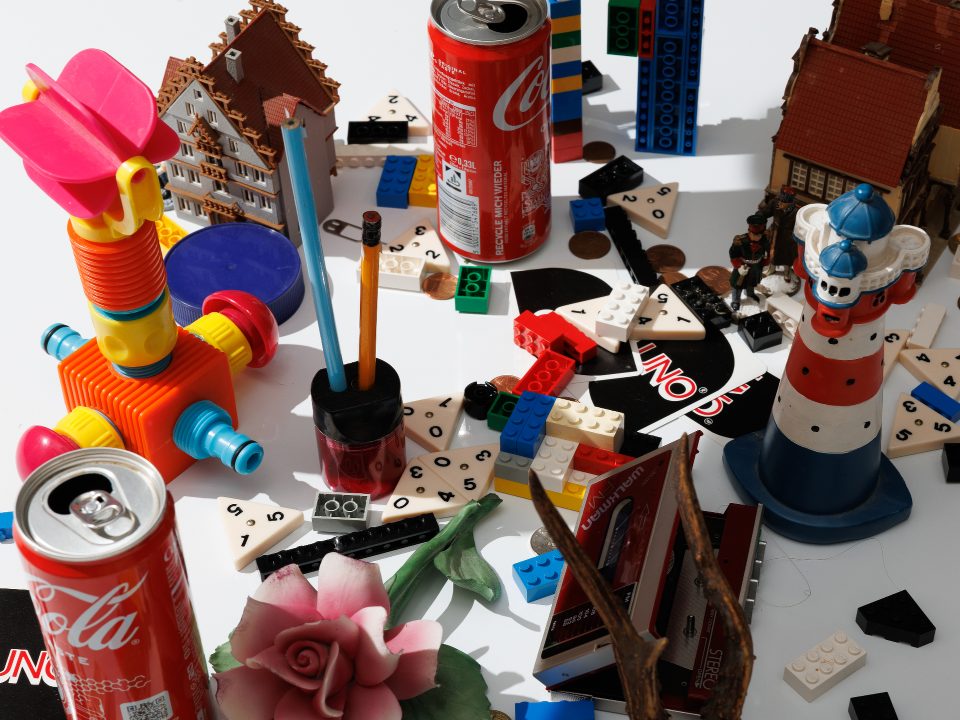} &
            \includegraphics[width=0.48\linewidth]{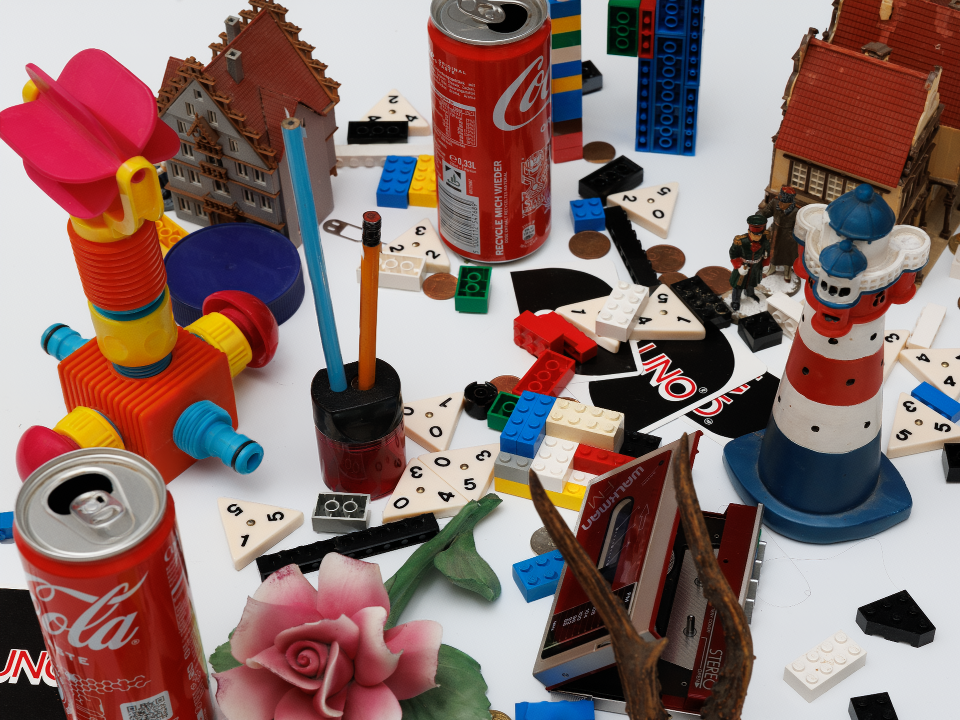}
        \end{tabular}
        \caption{Cluttered multi-object scene.}
    \end{subfigure}

    \caption{Representative qualitative results on the WSRD+ 2026 test set, showing the shadowed input and corresponding restored output from our final challenge model. Ground truth is unavailable at submission time.}
    \label{fig:WSRD2026_qualitative}
\end{figure}

Since the WSRD+ 2026 challenge test set does not provide ground truth during evaluation, we additionally show representative qualitative results of our final model on several test images. Figure~\ref{fig:WSRD2026_qualitative} compares the shadowed inputs with the corresponding restored outputs. The examples illustrate robust shadow removal across cluttered scenes, textured materials, and close-up object configurations.

\subsection{Qualitative analysis of iterative refinement}
\begin{figure}[t]
    \centering
    \includegraphics[width=\linewidth]{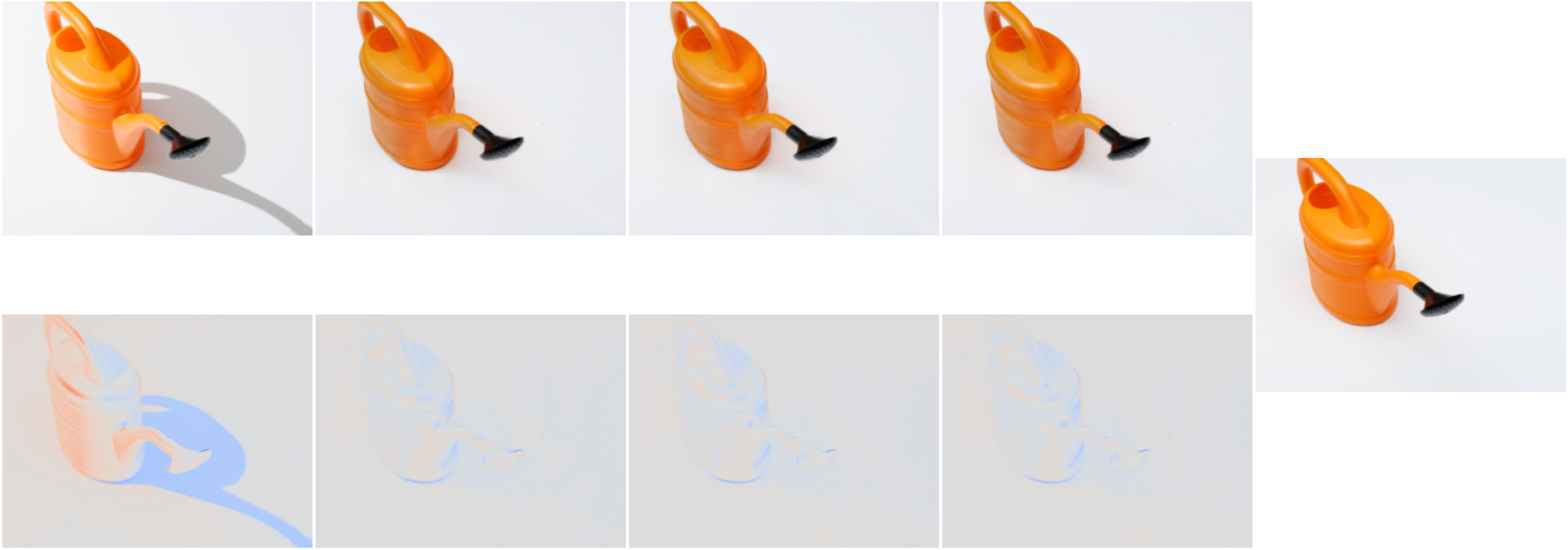}\\[1mm]
    \includegraphics[width=\linewidth]{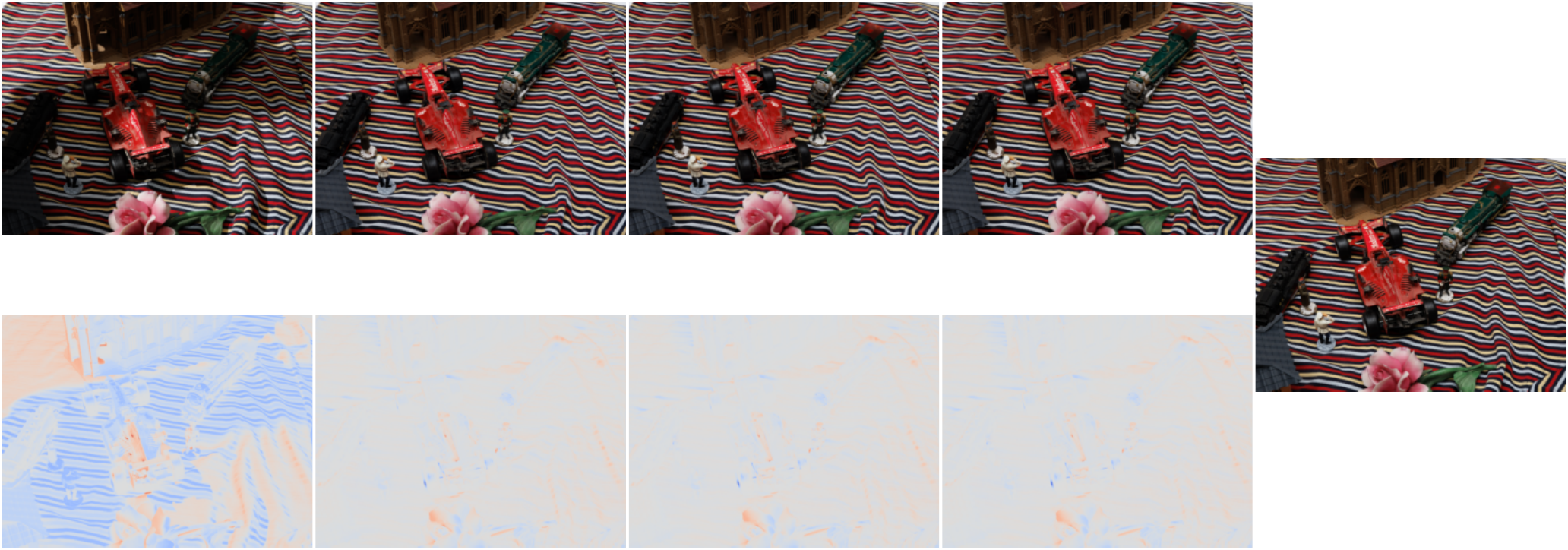}
    \caption{Stage-wise visualizations on the WSRD+ validation set. For each example, we show the RGB restoration strip and the corresponding difference with the ground truth. From left to right: the shadowed input, the outputs of Stages 1--3, and the ground truth. }
    \label{fig:iterative_process}
\end{figure}

We include representative qualitative examples to visualise the iterative shadow-removal process on the WSRD+ validation set. Figure~\ref{fig:iterative_process} shows the input image, the outputs of the three restoration stages, and the ground-truth target, together with the corresponding error maps. In both examples, the first stage performs the dominant shadow-removal correction, while the later stages mainly provide smaller residual refinements. The error maps further show a progressive reduction of residual artefacts across the cascade, especially around shadow boundaries and in low-frequency tonal regions.

\subsection{Results on ISTD+}

We report the results in \Cref{tab:istdplus_mask_usage_unified}.
On ISTD+, the pretrained model transfers poorly without adaptation.
After fine-tuning, performance improves substantially, indicating that domain adaptation is essential for this dataset, mainly due to the absence of self-cast shadows.
Our finetuned model achieves the best PSNR among the mask-free methods, surpassing the previous best result from OmniSR~\cite{xu2025omnisr} by \qty{0.90}{\deci\bel}, while remaining slightly below it in SSIM (\qty{-0.005}). The marginal difference in SSIM is consistent with our experiments in \Cref{tab:ablation_all}.
This suggests that our method improves overall reconstruction fidelity, though structural similarity is still marginally better captured by OmniSR.
Ensembling further improves performance to \qty{34.32} PSNR, but we regard this as an auxiliary result as it is not directly comparable to single-model baselines.
While our method remains below the strongest mask-conditioned models that use ground-truth shadow masks at inference time, it is competitive in PSNR. It outperforms several such baselines even operating without mask input.

\begin{table}[t]
\centering
\small
\setlength{\tabcolsep}{5pt}
\caption{ISTD+ whole-image results under the unified evaluation protocol of~\cite{xu2025detail}. Input mask denotes the mask used at evaluation time in that protocol. For mask-conditioned methods (b), the dataset-provided ISTD+ shadow mask is used as the default input. For group (a), no image mask is used. Evaluation is reported on the whole image at a resolution of $256\times256$ using PSNR/SSIM.}
\label{tab:istdplus_mask_usage_unified}
\begin{tabular}{llccc}
\toprule
 & Method  & PSNR$\uparrow$ & SSIM$\uparrow$ & LPIPS $\downarrow$ \\
\midrule
\multirow{7}{*}{(a)}
 & TBRNet~\cite{liu2023shadowbilinear}          & 31.91 & 0.964 & --  \\
 & Refusion~\cite{luo2023refusion}              & 32.41 & 0.961 & -- \\
 & DeS3~\cite{jin2024des3}                      & 31.39 & 0.957 & -- \\
 & OmniSR~\cite{xu2025omnisr}                   & 33.34 & \textbf{0.970} & -- \\
 & \textbf{Ours (pretrained only)}              & 18.44 & 0.895 & -- \\
 & \textbf{Ours (finetuned)}                    & 34.24 & 0.965 & 0.0235 \\
 & \textbf{Ours (ensemble top-5)}               & \textbf{34.32} & 0.965 & -- \\
\midrule
\multirow{7}{*}{(b)}
 & AutoExposure~\cite{fu2021auto}               & 29.45 & 0.861 & -- \\
 & BMNet~\cite{zhu2022bijective}               & 33.98 & 0.972 & -- \\
 & ShadowFormer~\cite{guo2023shadowformer}      & 35.46 & 0.971 & -- \\
 & DMTN~\cite{liu2023decoupled}                 & 32.23 & 0.966 & -- \\
 & ShadowDiffusion~\cite{guo2023shadowdiffusion} & 34.63 & 0.967 & -- \\
 & HomoFormer~\cite{xiao2024homoformer}         & \textbf{35.72} & \textbf{0.977} & -- \\
 & InstanceShadow~\cite{Mei2024latent}          & 34.69 & 0.968 & -- \\
\bottomrule
\end{tabular}
\end{table}

\subsection{Results on UAV-SC+}
We report the results in \Cref{tab:uavsc-whole}.
On UAV-SC+, the pretrained model transfers poorly without adaptation.
After fine-tuning, performance improves substantially, indicating that domain adaptation is essential for this dataset.
Our finetuned model achieves the best PSNR among all compared methods, surpassing the previous best result from RRMamba~\cite{chu2025rmmamba} by \qty{0.38} {\deci\bel}, while remaining slightly below it in SSIM (\qty{-0.015}).
This suggests that our method improves overall reconstruction fidelity, though structural similarity is still marginally better captured by RRMamba. Ensembling further improves performance to \qty{24.76}{\deci\bel} PSNR, but we regard this as an auxiliary result since it is not directly comparable to single-model baselines.

\begin{table}[t]
    \centering
    \small
    \setlength{\tabcolsep}{4pt}
    \caption{Whole-image quantitative results on the UAV-SC dataset.
    PSNR and SSIM for baseline methods are extracted from the all-image column of Table~II~\cite{chu2025rmmamba}.
    $\star$ denotes methods implemented by the source-paper authors, as code was not publicly available.}
    \label{tab:uavsc-whole}
    \begin{tabular}{@{}lccc@{}}
        \toprule
        Method & PSNR $\uparrow$ & SSIM $\uparrow$ & LPIPS $\downarrow$ \\
        \midrule
        Silva et al.~\cite{silva2018near}            & 16.26 & 0.687 & -- \\
        Gong et al.~\cite{gong2014interactive}              & 16.83 & 0.699 & -- \\
        Mask-ShadowGAN~\cite{hu2019mask}   & 19.12 & 0.736 & -- \\
        DC-ShadowNet~\cite{jin2021dc}     & 21.05 & 0.784 & -- \\
        LG-ShadowNet~\cite{liu2021shadow}       & 18.05 & 0.762 & -- \\
        G2R-ShadowNet~\cite{liu2021shadowgen}     & 16.83 & 0.718 & -- \\
        DMTN~\cite{liu2023decoupled}               & 23.00 & 0.818 & -- \\
        ST-CGAN~\cite{wang2018stcgan}$~\star$    & 23.40 & 0.833 & -- \\
        ShadowFormer~\cite{guo2023shadowformer}      & 23.10 & 0.829 & -- \\
        TBRNet~\cite{liu2023shadowbilinear}           & 22.50 & 0.770 & -- \\
        RASM~\cite{liu2024regional}             & 23.86 & 0.868 & -- \\
        RRMamba~\cite{chu2025rmmamba}             & 24.22 & \textbf{0.871} & -- \\
        
        \addlinespace
        \textbf{Ours (pretrained only)} & 15.45 & 0.695 & 0.4272 \\
        \textbf{Ours (finetuned)}       & 24.60 & 0.856 & 0.2298 \\
        \textbf{Ours (ensemble top-5)}  & \textbf{24.76} & 0.858 & \textbf{0.2253} \\
        \bottomrule
    \end{tabular}
\end{table}
\section{Conclusion}

We introduced a three-stage progressive shadow-removal framework based on OmniSR, where each stage explicitly refines the residual errors of the previous prediction using shared semantic guidance from DINOv2 and geometric cues derived from monocular depth and surface normals. The resulting system achieved first place on the NTIRE 2026 WSRD+ challenge, confirming that progressive direct refinement is an effective design for high-quality shadow removal under challenging illumination and distribution shift.

Importantly, the method also generalises beyond the challenge benchmark. After fine-tuning, it achieves the best PSNR among the compared mask-free methods on ISTD+, and the best PSNR overall on UAV-SC+. The UAV-SC+ result is particularly notable, since UAV imagery differs substantially from conventional shadow-removal benchmarks in viewpoint, scene composition, texture statistics, and shadow geometry. Strong performance in this setting suggests that the proposed model learns a transferable restoration prior rather than a dataset-specific solution.
At the same time, the transfer behaviour across datasets is not uniform. While ISTD+ remains an important benchmark, it typically contains comparatively simpler scenes and less geometric variability than WSRD+/WSRD+ or UAV-based imagery. We therefore hypothesise that the strong pretraining on WSRD+, which exposes the model to richer geometric structure, more complex textures, and more challenging illumination patterns, is less directly expressed on ISTD+ than on UAV-SC+. This may explain why the cross-dataset benefit appears especially compelling on the UAV benchmark, where the learned semantic and geometric priors are more aligned with the complexity of the target domain.

These results show that semantic and geometric guidance, when coupled with progressive multi-stage refinement, provide a strong foundation for robust deshadowing across heterogeneous domains. Future work will focus on improving the trade-off between distortion-oriented fidelity and perceptual quality, reducing the computational cost of multi-stage inference, and further investigating transfer to remote-sensing and other out-of-domain shadow-removal scenarios.

\section*{Acknowledgments}
This work was carried out within the SAFIR research project funded by the Austrian Research Promotion Agency (FFG) as part of the Research, Technology \& Innovation (RTI) initiative \enquote{Digitaler Zwilling Österreich}. This work was partially supported by the Alexander von Humboldt Foundation. Additionally, the computational results have been achieved using the Austrian Scientific Computing (ASC) infrastructure. Code and materials are available on GitHub: \url{https://github.com/AIT-Assistive-Autonomous-Systems/SGCR-SR.git}.

{
    \small
    \bibliographystyle{ieeenat_fullname}
    \bibliography{main}

@String(AAAI = {AAAI})

@String(BMVC= {Brit. Mach. Vis. Conf.})

@String(CVPR= {IEEE Conf. Comput. Vis. Pattern Recog.})

@String(ECCV= {Eur. Conf. Comput. Vis.})

@String(ICCV= {Int. Conf. Comput. Vis.})

@String(TIP  = {IEEE Trans. Image Process.})

@article{zhang2018improving,
  title={Improving shadow suppression for illumination robust face recognition},
  author={Zhang, Wuming and Zhao, Xi and Morvan, Jean-Marie and Chen, Liming},
  journal={IEEE Transactions on Pattern Analysis and Machine Intelligence},
  volume={41},
  number={3},
  pages={611--624},
  year={2018}
}

@article{cucchiara2003detecting,
  title={Detecting moving objects, ghosts, and shadows in video streams},
  author={Cucchiara, Rita and Grana, Costantino and Piccardi, Massimo and Prati, Andrea},
  journal={IEEE Transactions on Pattern Analysis and Machine Intelligence},
  volume={25},
  number={10},
  pages={1337--1342},
  year={2003}
}

@article{arif2022comprehensive,
  title={A comprehensive review of vehicle detection techniques under varying moving cast shadow conditions using computer vision and deep learning},
  author={Arif, Muhammad Umair and Farooq, Muhammad Umar and Raza, Rana Hammad and Lodhi, Zain Ul Abideen and Hashmi, Muhammad Abdur Rehman},
  journal={IEEE Access},
  volume={10},
  pages={104863--104886},
  year={2022}
}

@article{alavipanah2022shadow,
  title={The shadow effect on surface biophysical variables derived from remote sensing: a review},
  author={Alavipanah, Seyed Kazem and Karimi Firozjaei, Mohammad and Sedighi, Amir and Fathololoumi, Solmaz and Zare Naghadehi, Saeid and Saleh, Samiraalsadat and Naghdizadegan, Maryam and Gomeh, Zinat and Arsanjani, Jamal Jokar and Makki, Mohsen and others},
  journal={Land},
  volume={11},
  number={11},
  pages={2025},
  year={2022}
}

@article{inoue2020learning,
  title={Learning from synthetic shadows for shadow detection and removal},
  author={Inoue, Naoto and Yamasaki, Toshihiko},
  journal={IEEE Transactions on Circuits and Systems for Video Technology},
  volume={31},
  number={11},
  pages={4187--4197},
  year={2020}
}

@inproceedings{le2019shadow,
  title={Shadow removal via shadow image decomposition},
  author={Le, Hieu and Samaras, Dimitris},
  booktitle={ICCV},
  pages={8578--8587},
  year={2019}
}

@inproceedings{xu2025omnisr,
  title={OmniSR: Shadow Removal under Direct and Indirect Lighting},
  author={Xu, Jiamin and Li, Zelong and Zheng, Yuxin and Huang, Chenyu and Gu, Renshu and Xu, Weiwei and Xu, Gang},
  booktitle={AAAI},
  volume={39},
  number={8},
  pages={8887--8895},
  year={2025}
}

@article{oquab2023dinov2,
  title={DINOv2: Learning Robust Visual Features without Supervision},
  author={Oquab, Maxime and Darcet, Timothee and Moutakanni, Theo and Vo, Huy and Szafraniec, Marc and Khalidov, Vasil and Fernandez, Pierre and Haziza, Daniel and Massa, Francisco and El-Nouby, Alaa and Assran, Mahmoud and others},
  journal={arXiv preprint arXiv:2304.07193},
  year={2023}
}

@article{depth_anything_v2,
  title={Depth Anything V2},
  author={Yang, Lihe and Kang, Bingyi and Huang, Zilong and Zhao, Zhen and Xu, Xiaogang and Feng, Jiashi and Zhao, Hengshuang},
  journal={arXiv:2406.09414},
  year={2024}
}

@inproceedings{venkatakrishnan2013plug,
  title={Plug-and-Play Priors for Model Based Reconstruction},
  author={Venkatakrishnan, Singanallur V. and Bouman, Charles A. and Wohlberg, Brendt},
  booktitle={2013 IEEE Global Conference on Signal and Information Processing},
  pages={945--948},
  year={2013}
}

@article{romano2017red,
  title={The Little Engine That Could: Regularization by Denoising (RED)},
  author={Romano, Yaniv and Elad, Michael and Milanfar, Peyman},
  journal={SIAM Journal on Imaging Sciences},
  volume={10},
  number={4},
  pages={1804--1844},
  year={2017}
}

@inproceedings{xu2025detail,
  title={Detail-preserving latent diffusion for stable shadow removal},
  author={Xu, Jiamin and Zheng, Yuxin and Li, Zelong and Wang, Chi and Gu, Renshu and Xu, Weiwei and Xu, Gang},
  booktitle={Proceedings of the IEEE/CVF Conference on Computer Vision and Pattern Recognition},
  pages={7592--7602},
  year={2025}
}

@article{reehorst2019red,
  title={Regularization by Denoising: Clarifications and New Interpretations},
  author={Reehorst, Eric T. and Schniter, Philip},
  journal={IEEE Transactions on Computational Imaging},
  volume={5},
  number={1},
  pages={52--67},
  year={2019}
}

@inproceedings{vasluianu2024ntire,
  title={NTIRE 2024 image shadow removal challenge report},
  author={Vasluianu, Florin-Alexandru and Seizinger, Tim and Zhou, Zhuyun and Wu, Zongwei and Chen, Cailian and Timofte, Radu and Dong, Wei and Zhou, Han and Tian, Yuqiong and Chen, Jun and others},
  booktitle={Proceedings of the IEEE/CVF conference on computer vision and pattern recognition},
  pages={6547--6570},
  year={2024}
}

@INPROCEEDINGS{Mei2024latent,
  author={Mei, Kangfu and Figueroa, Luis and Lin, Zhe and Ding, Zhihong and Cohen, Scott and Patel, Vishal M.},
  booktitle={2024 IEEE/CVF Winter Conference on Applications of Computer Vision (WACV)}, 
  title={Latent Feature-Guided Diffusion Models for Shadow Removal}, 
  year={2024},
  volume={},
  number={},
  pages={4301-4310},
  doi={10.1109/WACV57701.2024.00426}}

@inproceedings{vasluianu2023wsrd,
  title={Wsrd: A novel benchmark for high resolution image shadow removal},
  author={Vasluianu, Florin-Alexandru and Seizinger, Tim and Timofte, Radu},
  booktitle={Proceedings of the IEEE/CVF Conference on Computer Vision and Pattern Recognition},
  pages={1826--1835},
  year={2023}
}

@inproceedings{ryu2019plug,
  title={Plug-and-Play Methods Provably Converge with Properly Trained Denoisers},
  author={Ryu, Ernest and Liu, Jialin and Wang, Sicheng and Chen, Xiaohan and Wang, Zhangyang and Yin, Wotao},
  booktitle={ICML},
  pages={5546--5557},
  year={2019}
}

@article{tan2025denoisers_inverse_problems,
  title={From Image Denoisers to Regularizing Imaging Inverse Problems: An Overview},
  author={Tan, Hong Ye and Mukherjee, Subhadip and Tang, Junqi},
  journal={arXiv preprint arXiv:2509.03475},
  year={2025}
}

@article{liu2021shadow,
  title={Shadow removal by a lightness-guided network with training on unpaired data},
  author={Liu, Zhihao and Yin, Hui and Mi, Yang and Pu, Mengyang and Wang, Song},
  journal={IEEE Transactions on Image Processing},
  volume={30},
  pages={1853--1865},
  year={2021},
  publisher={IEEE}
}

@article{liu2023decoupled,
  title={A decoupled multi-task network for shadow removal},
  author={Liu, Jiawei and Wang, Qiang and Fan, Huijie and Li, Wentao and Qu, Liangqiong and Tang, Yandong},
  journal={IEEE Transactions on Multimedia},
  volume={25},
  pages={9449--9463},
  year={2023},
  publisher={IEEE}
}

@inproceedings{liu2021shadowgen,
  title={From shadow generation to shadow removal},
  author={Liu, Zhihao and Yin, Hui and Wu, Xinyi and Wu, Zhenyao and Mi, Yang and Wang, Song},
  booktitle={Proceedings of the IEEE/CVF conference on computer vision and pattern recognition},
  pages={4927--4936},
  year={2021}
}

@inproceedings{jin2021dc,
  title={Dc-shadownet: Single-image hard and soft shadow removal using unsupervised domain-classifier guided network},
  author={Jin, Yeying and Sharma, Aashish and Tan, Robby T},
  booktitle={Proceedings of the IEEE/CVF international conference on computer vision},
  pages={5027--5036},
  year={2021}
}

@inproceedings{hu2019mask,
  title={Mask-shadowgan: Learning to remove shadows from unpaired data},
  author={Hu, Xiaowei and Jiang, Yitong and Fu, Chi-Wing and Heng, Pheng-Ann},
  booktitle={Proceedings of the IEEE/CVF international conference on computer vision},
  pages={2472--2481},
  year={2019}
}

@inproceedings{gong2014interactive,
  title={Interactive shadow removal and ground truth for variable scene categories},
  author={Gong, Han and Cosker, DP},
  booktitle={BMVC 2014-Proceedings of the British Machine Vision Conference 2014},
  year={2014}
}

@article{silva2018near,
  title={Near real-time shadow detection and removal in aerial motion imagery application},
  author={Silva, Guilherme F and Carneiro, Grace B and Doth, Ricardo and Amaral, Leonardo A and de Azevedo, Dario FG},
  journal={ISPRS Journal of photogrammetry and remote sensing},
  volume={140},
  pages={104--121},
  year={2018},
  publisher={Elsevier}
}

@article{chu2025rmmamba,
  title={RMMamba: Randomized Mamba for Remote Sensing Shadow Removal},
  author={Chu, Jun and Chi, Kaichen and Wang, Qi},
  journal={IEEE Transactions on Geoscience and Remote Sensing},
  year={2025},
  publisher={IEEE}
}

@inproceedings{jin2024des3,
  title={Des3: Adaptive attention-driven self and soft shadow removal using vit similarity},
  author={Jin, Yeying and Ye, Wei and Yang, Wenhan and Yuan, Yuan and Tan, Robby T},
  booktitle={Proceedings of the AAAI Conference on Artificial Intelligence},
  volume={38},
  number={3},
  pages={2634--2642},
  year={2024}
}

@inproceedings{ntire2026shadow,
  title={{NTIRE} 2026 Single Image Shadow Removal Challenge Report},
  author={Vasluianu, Florin-Alexandru and  Seizinger, Tim and  Zhou, Zhuyun and  WU, Zongwei and  Timofte, Radu
and others},
  booktitle={Proceedings of the IEEE/CVF Conference on Computer Vision and Pattern Recognition (CVPR) Workshops},
  year={2026}
}

@inproceedings{luo2023refusion,
  title={Refusion: Enabling large-size realistic image restoration with latent-space diffusion models},
  author={Luo, Ziwei and Gustafsson, Fredrik K and Zhao, Zheng and Sj{\"o}lund, Jens and Sch{\"o}n, Thomas B},
  booktitle={Proceedings of the IEEE/CVF conference on computer vision and pattern recognition},
  pages={1680--1691},
  year={2023}
}

@InProceedings{Vasluianu_2023_CVPR,
  author    = {Vasluianu, Florin-Alexandru and Seizinger, Tim and Timofte, Radu },
  title     = {NTIRE 2023 Image Shadow Removal Challenge Report},
  booktitle = {Proceedings of the IEEE/CVF Conference on Computer Vision and Pattern Recognition (CVPR) Workshops},
  month     = {June},
  year      = {2023},
  pages     = {1788--1807}
}

@inproceedings{zhu2022bmnet,
  title={Bidirectional Residual Transformer for Shadow Removal},
  author={Zhu, Yurui and Xiao, Jie and Fu, Xueyang and others},
  booktitle={ECCV},
  year={2022}
}

@inproceedings{liu2024regional,
  title={Regional attention for shadow removal},
  author={Liu, Hengxing and Li, Mingjia and Guo, Xiaojie},
  booktitle={Proceedings of the 32nd ACM International Conference on Multimedia},
  pages={5949--5957},
  year={2024}
}

@article{liu2023shadowbilinear,
  title={A shadow imaging bilinear model and three-branch residual network for shadow removal},
  author={Liu, Jiawei and Wang, Qiang and Fan, Huijie and Tian, Jiandong and Tang, Yandong},
  journal={IEEE Transactions on Neural Networks and Learning Systems},
  volume={35},
  number={11},
  pages={15857--15871},
  year={2023},
  publisher={IEEE}
}

@inproceedings{fu2021auto,
  title={Auto-exposure fusion for single-image shadow removal},
  author={Fu, Lan and Zhou, Changqing and Guo, Qing and Juefei-Xu, Felix and Yu, Hongkai and Feng, Wei and Liu, Yang and Wang, Song},
  booktitle={Proceedings of the IEEE/CVF conference on computer vision and pattern recognition},
  pages={10571--10580},
  year={2021}
}

@inproceedings{zhu2022bijective,
  title={Bijective mapping network for shadow removal},
  author={Zhu, Yurui and Huang, Jie and Fu, Xueyang and Zhao, Feng and Sun, Qibin and Zha, Zheng-Jun},
  booktitle={Proceedings of the IEEE/CVF conference on computer vision and pattern recognition},
  pages={5627--5636},
  year={2022}
}

@inproceedings{guo2023shadowformer,
  title={ShadowFormer: Global Context Helps Image Shadow Removal},
  author={Guo, Lanqing and Huang, Siyu and Liu, Ding and Cheng, Hao and Wen, Bihan},
  booktitle={AAAI},
  year={2023}
}

@article{luo2023evolutionary,
  title={An evolutionary shadow correction network and a benchmark UAV dataset for remote sensing images},
  author={Luo, Shuang and Li, Huifang and Li, Yiqiu and Shao, Chenglin and Shen, Huanfeng and Zhang, Liangpei},
  journal={IEEE Transactions on Geoscience and Remote Sensing},
  volume={61},
  pages={1--14},
  year={2023},
  publisher={IEEE}
}

@inproceedings{wang2018stcgan,
  title={Stacked conditional generative adversarial networks for jointly learning shadow detection and shadow removal},
  author={Wang, Jifeng and Li, Xiang and Yang, Jian},
  booktitle={Proceedings of the IEEE conference on computer vision and pattern recognition},
  pages={1788--1797},
  year={2018}
}

@inproceedings{xiao2024homoformer,
  title={HomoFormer: Homogenized Transformer for Image Shadow Removal},
  author={Xiao, Jie and Fu, Xueyang and Zhu, Yurui and Li, Dong and Huang, Jie and Zhu, Kai and Zha, Zheng-Jun},
  booktitle={CVPR},
  pages={25617--25626},
  year={2024}
}

@inproceedings{guo2023shadowdiffusion,
  title={ShadowDiffusion: When Degradation Prior Meets Diffusion Model for Shadow Removal},
  author={Guo, Lanqing and Wang, Chong and Yang, Wenhan and Huang, Siyu and Wang, Yufei and Pfister, Hanspeter and Wen, Bihan},
  booktitle={CVPR},
  pages={14049--14058},
  year={2023}
}

@InProceedings{zhang2018lpips,
  author    = {Zhang, Richard and Isola, Phillip and Efros, Alexei A. and Shechtman, Eli and Wang, Oliver},
  booktitle = {IEEE/CVF Conference on Computer Vision and Pattern Recognition},
  title     = {The Unreasonable Effectiveness of Deep Features as a Perceptual Metric},
  year      = {2018},
  pages     = {586--595},
  doi       = {10.1109/cvpr.2018.00068},
}

@Article{Wang04:IQA,
  author  = {Zhou Wang and Bovik, Alan C. and Sheikh, Hamid R. and Simoncelli, Eero P.},
  journal = {IEEE Transactions on Image Processing (TIP)},
  title   = {Image Quality Assessment: From Error Visibility toStructural Similarity},
  year    = {2004},
  number  = {4},
  pages   = {600--612},
  volume  = {13},
  doi     = {10.1109/tip.2003.819861},
}

@InProceedings{Jayasumana24:FID,
  author    = {Jayasumana, Sadeep and Ramalingam, Srikumar and Veit, Andreas and Glasner, Daniel and Chakrabarti, Ayan and Kumar, Sanjiv},
  booktitle = {IEEE/CVF Conference on Computer Vision and Pattern Recognition (CVPR)},
  title     = {Rethinking FID: Towards a Better Evaluation Metric for Image Generation},
  year      = {2024},
  month     = jun,
  pages     = {9307--9315},
  publisher = {IEEE},
  doi       = {10.1109/cvpr52733.2024.00889},
}
}
\end{document}